\documentclass[review]{elsarticle}

\usepackage{hyperref}
\usepackage{array}
\usepackage{hhline}
\usepackage{makecell}
\usepackage{subfig}
\usepackage{color}
\usepackage[T1]{fontenc}
\usepackage{amsmath}

\newcolumntype{M}[1]{>{\centering}m{#1}}
\journal{Pattern Recognition}

\begin{document}

\begin{frontmatter}

%\title{Cohort of LSTM and lexicon verification for handwriting recognition with gigantic lexicon}
\title{Handwriting recognition using Cohort of LSTM and lexicon verification with extremely large lexicon}

%% Group authors per affiliation:
%% \author{Elsevier\fnref{myfootnote}}
%% \address{Radarweg 29, Amsterdam}

%% or include affiliations in footnotes:
\author[univ]{Bruno STUNER\corref{mycorrespondingauthor}}
\cortext[mycorrespondingauthor]{Corresponding author}
\ead{bruno.stuner@litislab.eu}

\author[univ]{Cl\'ement CHATELAIN}
\ead{clement.chatelain@litislab.eu}

\author[univ]{Thierry PAQUET}
\ead{thierry.paquet@litislab.eu}

\address[univ]{Normandie Univ, UNIROUEN, UNIHAVRE, INSA Rouen, LITIS, 76000 Rouen, France}

\begin{abstract}
State-of-the-art methods for handwriting recognition are based on Long Short Term Memory (LSTM) recurrent neural networks (RNN), which now provides very impressive character recognition performance. 
The character recognition is generally coupled with a lexicon driven decoding process which integrates dictionaries. 
Unfortunately these dictionaries are limited to hundred of thousands words for the best systems, which prevent from having a good language coverage, and therefore limit the global recognition performance. 
In this article, we propose an alternative to the lexicon driven decoding process based on a \emph{lexicon verification} process, coupled with an original cascade architecture. 
The cascade is made of a large number of complementary networks extracted from a single training (called cohort), making the learning process very light. 
The proposed method achieves new state-of-the art word recognition performance on the Rimes and IAM datasets. 
Dealing with gigantic lexicon of 3 millions words, the methods also demonstrates interesting performance with a fast decision stage.

%Handwriting recognition state of the art methods are based on Long Short Term Memory (LSTM) recurrent neural networks (RNN) coupled with the use of linguistic knowledge. 
%LSTM RNN presents high raw performance and interesting training properties that allow us to break with the standard method at the state of the art.
%We present a simple and efficient way to extract from a single training a large number of complementary LSTM RNN, called cohort, combined in a cascade architecture with a lexical verification.
%This process does not require fine tuning, making it easy to use.
%Our verification allows to deal quickly and efficiently with gigantic lexicon (over 3 million words).
%We achieve state of the art results for isolated word recognition with very large lexicon and present novel results for an unprecedented gigantic lexicon.
\end{abstract}


\end{frontmatter}

\section{Introduction}

Handwriting recognition is the numeric process of translating handwritten text images into strings of characters.
The handwriting recognition process traditionally involves two steps \cite{plamondon2000online}: optical character recognition and linguistic processing. 
Optical character recognition is a hard task due to the variability of shapes in handwritten texts, since every human has his own personal writing style.
Therefore, even when using state of the art classifiers like deep neural networks to recognize characters \cite{graves2009offline}, a considerable amount of errors would occur by considering only the optical model.
%Indeed optical models are intended to provide character classes likelihoods from the observation of the image only.
Linguistic processing aims at combining the characters hypotheses together so as to provide the most likely sequence of words in accordance with some high level linguistic rules.
There are two types of linguistic knowledge: lexicons and language models. A language model is a probabilistic modelization of a language which generally provides word sequence likelihood, allowing to rank the recognition hypothesis provided by the optical model.
Nowadays, the use of linguistic knowledge is an open problem.
Lexicon driven approaches aim at recognizing words thanks to the use of a lexicon. 
They search for the most likely word that belongs to the working lexicon, by concatenating the character hypotheses.
There is currently no efficient alternative to the use of lexicon driven recognition methods either for isolated words or for text recognition.
Which lexicon resource should be used, which corpora should be selected for training the language model ?
These choices directly affect the recognition performance.
In the case of lexicon driven methods, where the characters are aligned on the lexicon words, too small lexicons fail to cover the test dataset, thus missing solutions. 
However, using a large lexicon (1000 words and more) requires many computations and generally produces precision loss \cite{koerich2003large}.
%Clem : j'ai supprimé le paragraphe suivant car les méthodes sont discutées après.
%To tackle these problems, many solutions exist like lexicon pruning, which enable to reduce lexicon size, or using n-grams of characters to deal with out of vocabulary words. However they hardly reach the state of the art performance.
To the best of our knowledge, the largest lexicon used in the literature was composed of 200K words \cite{hamdani2014rwth} (60K words for \cite{bluche2014a2ia}), and there could still occur out of vocabulary words (named entity, numbers, etc).

During the last years, significant progress in handwriting recognition and especially in optical models have been made thanks to deep learning advances \cite{lecun2015deep}, namely with the Long Short Term Memories (LSTM) Recurrent Neural Networks (RNN) \cite{hochreiter1997long}.
The LSTM recurrent neural networks achieve state of the art performance in various applications involving sequence recognition, such as speech recognition \cite{graves2005framewise}, protein predictions \cite{thireou2007bidirectional}, machine translation \cite{sutskever2014sequence}, and optical character recognition \cite{grosicki2009icdar,menasri2012a2ia}.
%The Connectionist Temporal Classification (CTC) \cite{CTC} training framework allows frame level optimization while only providing the ground truth at word or line level.
%Therefore, it avoids a tedious manual segmentation of the images to provide the frame level ground truth.
Performance of complete systems including both LSTM networks and linguistic resources \cite{pham2014dropout} is due to the very high raw performance of the optical model (i.e. without using additional linguistic resource).
For example, the raw performance of the optical model is about 35\% WER on the RIMES dataset when using a BLSTM optical model solely. 
The contribution of the language model is then to penalize the wrong hypotheses produced by the optical model, so as to favor the most likely word sequences from the language model point of view.
We believe that the raw performance of the LSTM based optical models provide hints that such networks should be used in a more specific way, and not only as a character classifier using a lexicon directed recognition approach, as it is the case in most of the actual studies reported in the literature.

Breaking the standard use of LSTM RNN as a simple classifier introduced in a lexicon driven decoding scheme, this work proposes a new recognition paradigm that improves handwriting recognition state of the art performance.
This new paradigm is based on word classifiers combination using an efficient decision rule operating at word level, which consists in lexicon verification.
Lexicon verification consists in accepting a word recognition hypothesis if it belongs to the lexicon, and rejecting it otherwise.
The underlying idea is that it is very unlikely that a wrong word recognition hypothesis belongs to the lexicon. 
The major advantage of this strategy is that it constitutes an extremely fast decision process, especially when compared to the tedious lexicon-driven decoding process which generally consists in a Viterbi beam search \cite{fissore1988strategies}. 
Classifier combination is introduced using a cascade framework for combining multiple word classifiers. 
It is based on two key points: i) a lexicon verification decision process ii) a pool of complementary recognizers. 
We introduce a very efficient way to produce hundreds of complementary word recognizers in a very reasonable training time. 
Following the recent theoretical results in deep learning\cite{choromanska2015loss}, we observed that multiple complementary networks can be obtained during a single training stage. 
We exploit this theoretical result to produce hundreds of complementary LSTM networks using a single training.
%Instead of looking for the best network, as it is traditionally done, we obtain an ensemble of complementary networks which we call a cohort of LSTM networks.
%These complementary networks can efficiently be combined in a cascade\cite{viola2001rapid}. 
%Relying on the raw performances of each LSTM RNN, we show that recognition and rejection in the cascade can be implemented using a simple and fast lexicon verification strategy. 
%This provides a new lexicon verification decoding paradigm for handwriting recognition, in opposition to the standard lexicon driven decoding scheme.
We show that the proposed strategy reaches very high performance regardless the size of the lexicon.
As a consequence, the approach has no limitation regarding the lexicon size, as demonstrated by the results obtained using a gigantic lexicon of more than 3 million words.

This article starts by a review of the state of the art of handwriting recognition highlighting the latest results obtained with BLSTM networks and Hidden Markov Models.
In the second part, we present our approach made of a cascade of LSTM recurrent neural networks.
We show how to get complementary LSTM RNN during training.
In the third part of the paper, the implementation of the method is described.
Finally the results are presented on the Rimes and IAM datasets, and then discussed before concluding.

\section{Related works}

\subsection{Handwriting recognition}

\begin{figure}
\centering
\includegraphics[width=12cm]{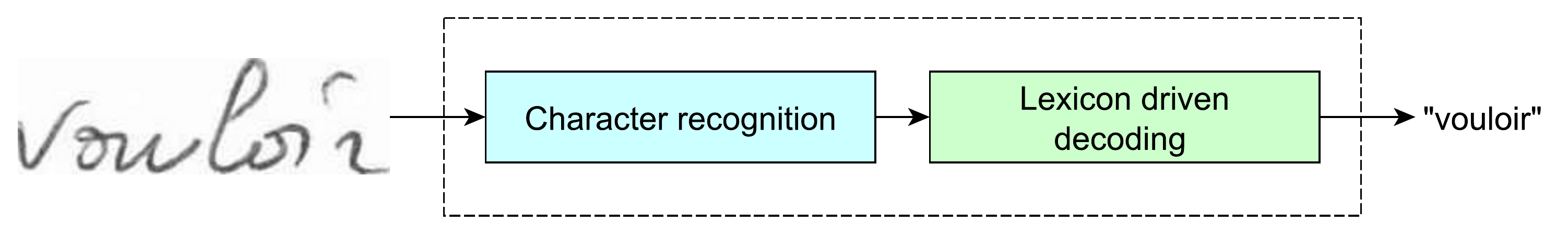}
\caption{The standard handwriting recognition paradigm.}
\label{HWRprocs1}
\end{figure}

Handwriting recognition models can be classified according to the character segmentation approach which can be either explicit (an algorithm specifically segments characters prior to their recognition), or implicit (characters are classified without prior segmentation)\cite{plamondon2000online}. 
Handwriting models can also be classified according to the character recognition method (discriminant classifiers for hybrid approaches \cite{senior1996forward} or generative approaches for Hidden Markov Models (HMM) \cite{el1999hmm}).
The common point of these approaches is that they all rely on a lexicon driven decoding stage.
Indeed, since character recognition is not perfect, the efficient word recognition strategy is to postpone the character decision process until the end of the sequence recognition process, where the best character sequence hypothesis being a valid sequence is finally selected. 
This traditional scheme is represented in Figure \ref{HWRprocs1}.

In the literature, isolated handwritten word recognition is the process of recognizing words that belong to a known lexicon which defines a closed vocabulary setting. 
The lexicon driven decoding methods used in this process are mainly based on Markov Models \cite{plotz2009markov,hamdani2013open,kozielski2013open,poznanski2016cnn}.
In this context, every well performing method, even those which use BLSTM networks, are based on lexicon driven decoding using either a lexicon or a character language model.
Their main strength is the segmentation free design, thus letting the decoding process provides the segmentation.

\subsection{The large vocabulary problem}

When using lexicon driven decoding, the character classification decisions are postponed until the end of the word so as to decide of the most probable word belonging to the working lexicon. 
On the one hand, lexicon directed approaches allow to correct character recognition errors. 
But on the other hand, they require using large lexicons in order to get high word coverage rates, and minimize Out Of Vocabulary (OOV) words.  
However, using very large lexicons requires pruning during the recognition phase, in order to get acceptable processing time.
In \cite{grosicki2009icdar} the authors relate the results of the RIMES 2009 competition for isolated word recognition, in which most of the participants have implemented HMM based approaches for which we can observe a neat difference of the recognition rate between small, medium and large lexicon size with a maximum size of 5334 words.
%At this time, only the TUM system was implementing BLSTM RNN which was less sensitive to lexicon effects.
The lexicon's size has also a major role in the use of these methods in real applications where the lexicon is linked to a language that can contain hundreds of thousands of words.
In specific cases, using such linguistic resources to improve recognition is not anymore compatible with real time constraints.

The large vocabulary problem is well known, as described in \cite{koerich2003large}, where a vocabulary of 1000 words is considered to be a large lexicon. 
Today larger lexicons are considered.
In \cite{pham2014dropout} the authors used 50k, 12k and 95k word lexicons for the IAM, Rimes and OpenHaRT dataset respectively, but none of these lexicons are fully covering the evaluation set.
To the best of our knowledge, the largest lexicon ever used is composed of 200 000 words \cite{hamdani2014rwth} (60K words for \cite{bluche2014a2ia}) using a n-gram model.
But this is still generating problems as out of vocabulary words still remain.
Moreover general purpose applications require unrestricted lexicons
which may be composed of hundreds of thousands words (e.g. French Gutenberg dictionary has 336k words) to reach an acceptable coverage rate, but still without covering named entities, numbers, etc. 

To overcome these limitations, solutions have been investigated: Pruning \cite{madhvanath1996holistic,madhvanath2001syntactic,gilloux1998reduction}, lexicon free decoding using character language models as an alternate solution to using a lexicon \cite{brakensiek2002handwritten, bharath2012hmm,shridhar1997handwritten,Cha06e,Cha06c}, sub lexical units driven recognition \cite{brakensiek2002handwritten,hamdani2013open,poznanski2016cnn}.
Pruning methods exhibit problems such as excessive pruning, leading to increasing errors.
Some methods\cite{brakensiek2002handwritten, bharath2012hmm} using hidden Markov models with other techniques like bag of symbols allow a lexicon free decoding , but they still have not reached the performance of the lexicon driven approaches.
Note that such approaches have also been used for numerical field recognition without lexicon \cite{shridhar1997handwritten,Cha06e,Cha06c}, but digits are simpler to recognize than characters since they are generally isolated, and there are only 10 classes.
Finally, in \cite{brakensiek2002handwritten,hamdani2013open,kozielski2013open,poznanski2016cnn} the authors proposed a lexicon decomposition into prefix and suffix of the word's lexicon, modeled by n-grams.
These methods are based on statistics extracted from a training corpus.
The n-grams models reach state of the art performance when dealing with out of lexicon word \cite{hamdani2013open,kozielski2013open}.

As reviewed in this paragraph, the very large vocabulary problem is still an open question.
With the current fast progress in deep learning, many architectures are studied. 
Regarding handwriting and speech recognition, it is currently lead by the Long Short Term Memory recurrent neural networks.

\subsection{Recurrent Neural Networks}
\label{sec:lstm}
Recurrent neural Networks (RNN), proposed more than 30 years ago with Hopfield networks \cite{hopfield1982neural}, get their efficiency from their ability to process sequences, thanks to recurrent connections bringing information about the previous inputs or states in the sequence to the current position.
However, for a long period of time, training recurrent neural networks suffered from the vanishing gradient problem \cite{hochreiter1998vanishing}.
As a consequence, long term dependencies can not be learned.
Long Short Term Memory (LSTM) cells have thus been designed by Hochreiter et al. \cite{hochreiter1997long} in order to overtake this limitation. Many improvements still have been proposed on LSTM \cite{gers2000learning,gers2003learning} adding a gate and peepholes.% and more recently on a stable LSTM for multidimensional RNN \cite{leifert2016cells}. 
Recently Gated Recurrent Units \cite{chung2014empirical} have been proposed as an alternative to LSTM units, requiring less computation but with almost similar performance. 
In this paper we focus on LSTM cells.

LSTM cells enable to learn long or short term dependencies while processing sequences thanks to the introduction of gates (input, forget and output gates) followed by a sigmoid function, which controls the internal memory cell update (updating, resetting, expressing) by introducing a multiplier cell at each gate.
RNN operate by processing the sequence in  a particular direction, that is why bi-directionality has been introduced in RNN \cite{schuster1997bidirectional}.
Then this idea has been extended to LSTM networks \cite{graves2005framewise} to create bidirectional LSTM recurrent neural networks (BLSTM). 
However key applications like handwriting recognition are based on images which have two dimensions.
In order to process images, multidimensional LSTM recurrent neural networks (MDLSTM) have been introduced \cite{graves2009offline}.

LSTM networks only became popular once a new learning strategy was introduced, the Connectionist Temporal Classification (CTC)\cite{CTC}, which is a neural variant of the well known Forward Backward algorithm used for training HMM.
CTC greatly helps training such networks by allowing embedded training of character models from the word or sentence label without the need for knowing the location of each character.
This is working especially well thanks to the introduction of a "joker" class between characters, which allows the network to postpone the decision until sufficient information is gathered along the sequence, so as to output the character hypothesis at one particular position in the input stream.

\begin{figure}[ht]
\centering
\includegraphics[width=8cm]{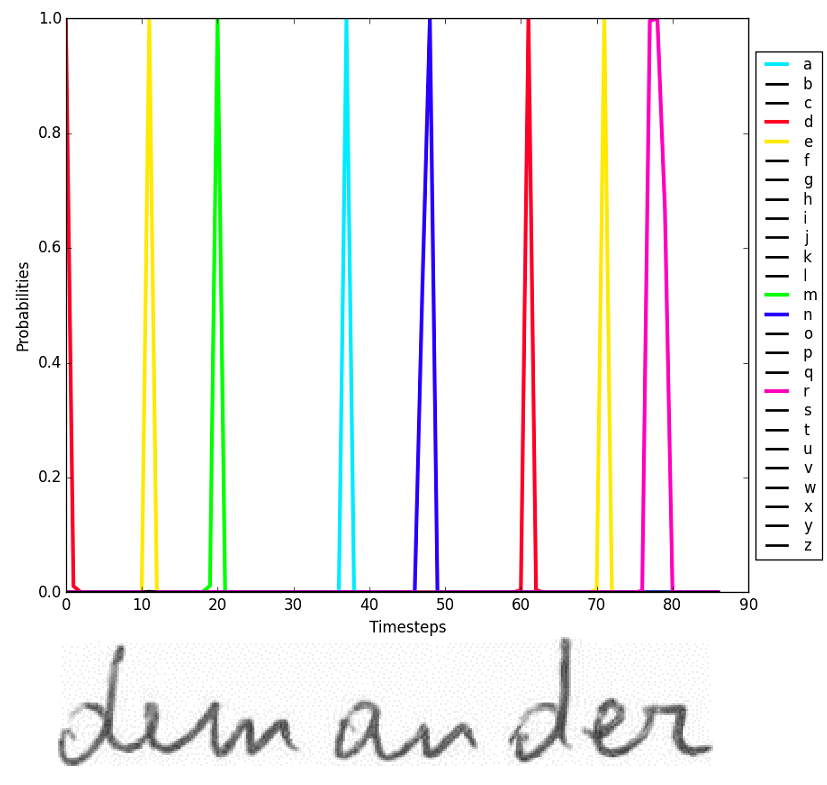}
\caption{LSTM network outputs at the end of a CTC training. 
The outputs form peaks for every character recognized.}
\label{ctcpeak}
\end{figure}

Today, LSTM recurrent neural networks are the state of the art method for numerous sequence analysis applications, including optical character recognition \cite{grosicki2009icdar,el2009icdar}.
LSTM networks provide nearly binary posterior probabilities.
One example of LSTM network outputs for the French word "demander" is given in Figure \ref{ctcpeak}, for readability of the legend, only lowercase characters are shown, with the correct characters highlighted with colors.
Such an output profile allows to apply a simple decoding scheme without lexicon, known as "Best path decoding" \cite{graves2012supervised}. 
It takes the maximum a posteriori probability of the character class at each frame, and by removing every successive repetitions of each class (joker included), then the joker.
The raw performance at the end of this lexicon free decoding scheme on a recognition task, although below state of the art, are very high. For example, on the Rimes word isolated recognition task, standard MDLSTM-RNN \cite{pham2014dropout} recognize 67\% of the words, by using this simple best path decoding strategy.

The LSTM RNN high performance shows that contribution to the recognition mostly comes from LSTM RNN (see Fig. \ref{HWRproc}).
From these observations and considering the aforementioned problems inherent to the use of lexicon driven recognition, in this paper we explore a new paradigm for isolated handwritten word recognition. 
The following section is devoted to the presentation of this new strategy which is based on combining multiple character classifiers using a word lexicon verification rule of the sequence's hypotheses.

\section{Proposed Approach}
\label{sec:proposedapproach}
\begin{figure}
\centering
\subfloat[The standard handwriting recognition paradigm]{\label{HWRproc}\includegraphics[width=12cm]{HWRprocess2.pdf}}
\hspace{5pt}
\subfloat[Our new handwriting recognition paradigm]{\label{HWRprocO}\includegraphics[width=12cm]{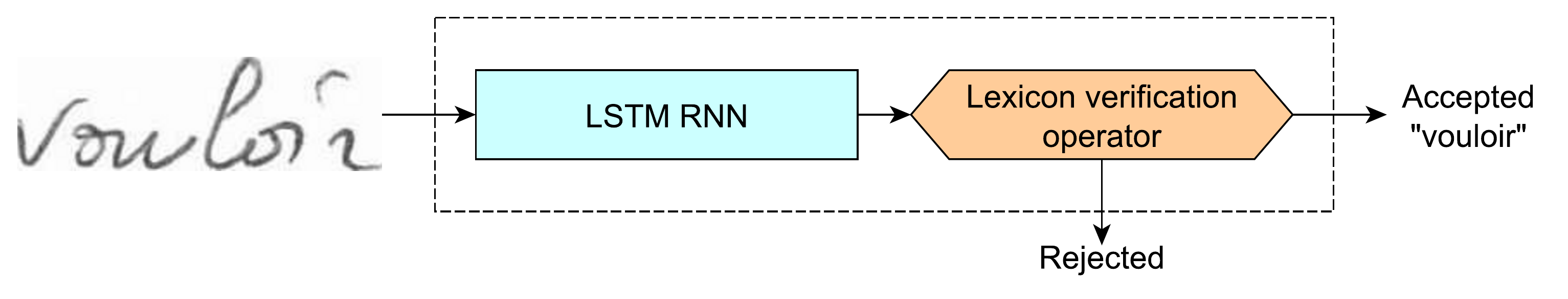}}
\caption{Handwriting recognition paradigm.}
\end{figure}

%In this paper, we propose a new recognition paradigm that takes benefit from the high raw performance of LSTM networks in order to carry out a lexicon verification decoding strategy, and suppress lexicon driven decoding.
This paper proposes a new recognition paradigm which provides an efficient alternative to lexicon driven decoding.
This strategy is based on cascading complementary neural networks, combined with a reliable rejection stage based on lexicon verification (see Figure \ref{HWRprocO}).
This section describes the overall cascade architecture (Section \ref{sec:cascade}), the proposed rejection stage (Section \ref{sec:verif}) and how complementary classifiers are generated (Section \ref{sec:train})

\subsection{The cascade framework}
\label{sec:cascade}

The first element of our lexicon driven decoding free approach is the cascade framework.
Cascade of classifiers is a combination method that sequentially combines classifiers decisions by exploiting the complementary behavior of the classifiers, in order to progressively refine recognition decisions along the cascade. 
The most famous contribution on cascade of classifiers is from Viola and Jones \cite{viola2001rapid}, who applied their framework to face detection.
The authors present a cascade based on a large ensemble of diverse and weak classifiers, which enable a fast decision process by sequentially introducing reliable decision stages.
The performance of each classifier may be low but it must be associated to a high confidence decision stage that accepts or rejects the decision.
In \cite{zhang2013reliable} and \cite{zhang2007novel}, an alternative cascade scheme is proposed by combining strong classifiers with different architecture and different input features.
The strong classifier recognizes an important number of objects with a low error rate, while relying on a decision stage allowing to transfer rejects to the next classifiers.

\begin{figure}[ht!]
\centering
\includegraphics[width=12cm]{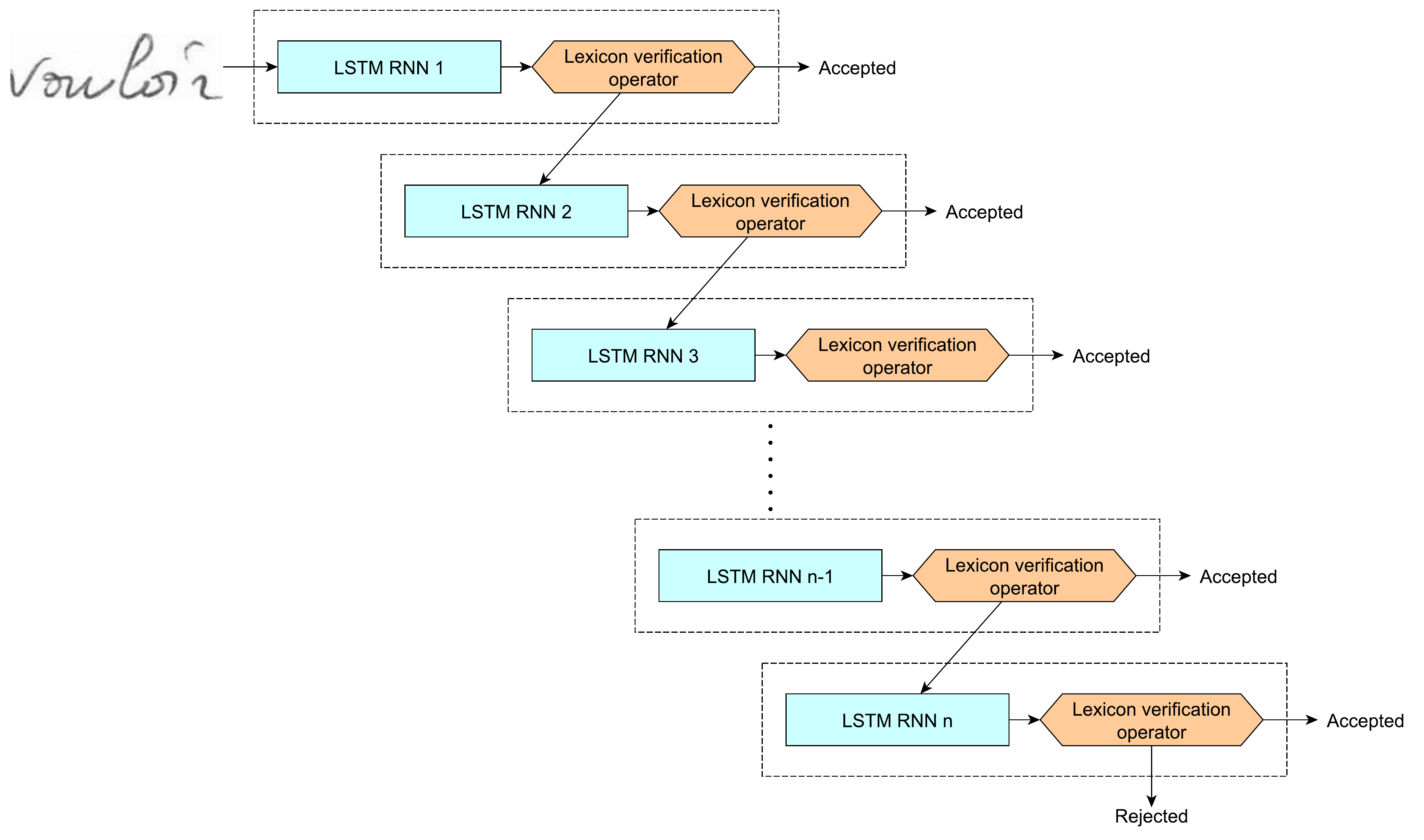}
\caption{The proposed BLSTM Cascade. Each classifier process the rejects from the previous layer.}
\label{schemaCascade}
\end{figure}

This brief literature review shows that for both strategies, using either strong or weak classifiers, the strength of a cascade scheme comes from the reliability of the decision stage introduced after each classifier.
In this paper we use the strong classifier approach, where we combine hundreds of LSTM RNN. 
Classifiers are ordered by increasing error (i.e. by decreasing reliability) and a reliable decision stage is introduced based on a lexicon verification operator (see Figure \ref{schemaCascade}).
When the lexicon verification operator accepts an hypothesis, the classification process stops thus avoiding to use the whole set of classifiers most of the time.
This decision mechanism is essential to control the performance of the cascade while enabling to significantly speed up the process, when many classifiers are involved. 

The second important element of a cascade is the classifiers complementarity.
Indeed, classifiers with similar behavior would not allow to refine the decision along the cascade.
%%%%%%%%%%%%%%%%%%%%
%%% CLEM : J'ai supprimé cette mini biblio pour alléger, car on revient dessus en détail dans la section 3.3. 
%%%%%%%%%%%%%%%%%%
%Complementary classifiers can be obtained following two different strategies. 
%The first one is by using different feature sets such as in \cite{zhang2007novel}, however designing many classifiers with different features is tedious and not straightforward.
%The second one is by selecting the training set of each classifiers in the cascade. 
%The popular scheme is based on training a classifier on the rejected samples of the previous stage in the cascade. 
%But this requires to have very large training datasets, as the amount of samples reduces at each stage of the cascade.
In this paper, we take benefit of the deep neural networks properties to get complementary classifiers during training of a single RNN, as discussed in section \ref{sec:train}. 

%%%%%%%%%%%%%%
% CLEM : idem, j'ai supprimé car redite par rapport à la section 3.3.
%%%%%%%%%%%%%%%
%Indeed, theoretical results regarding the loss surface of a neural network show that the many local minimums are of similar magnitude, and therefore provide multiple parameterization of the network with similar overall performance but with local differences \cite{choromanska2015loss}. We think that exploiting these many local minimums can provide complementary decision at the word level. The complementary classifier generation is discuss

We now discuss the decision reliability of our lexicon verification operator.

\subsection{Lexicon verification operator reliability}
\label{sec:verif}
The decision stage is made of a lexicon verification rule which consists in accepting a character string hypothesis if it belongs to the lexicon, or rejecting it otherwise.
Such simple verification stage can serve as a decision in the cascade only if it provides reliable decisions and has a very low false acceptance rate.

A False Acceptance (FA) of our lexicon verification rule occurs when a wrong character string hypothesis produced by the recognizer matches one entry of the lexicon.  
The probability of false acceptance $P_{FA}$ for a given recognizer can be expressed as described in equation (1) where $W$ is a word hypothesis, $L$ a lexicon, and $Reco$ a recognizer.
"$W\text{ \textbf{is} erroneous}$" is $true$ when the word hypothesis $W$ is wrong, and "$W\in{} L$" is $true$ when the word hypothesis $W$ belongs to the lexicon. 

\begin{equation}
P_{FA}   =  P(W\text{ \textbf{is}~erroneous~} \wedge{} W \in{} L \mid{}Reco)   
\end{equation}

\begin{figure}[ht]
\centering
\includegraphics[width=10cm]{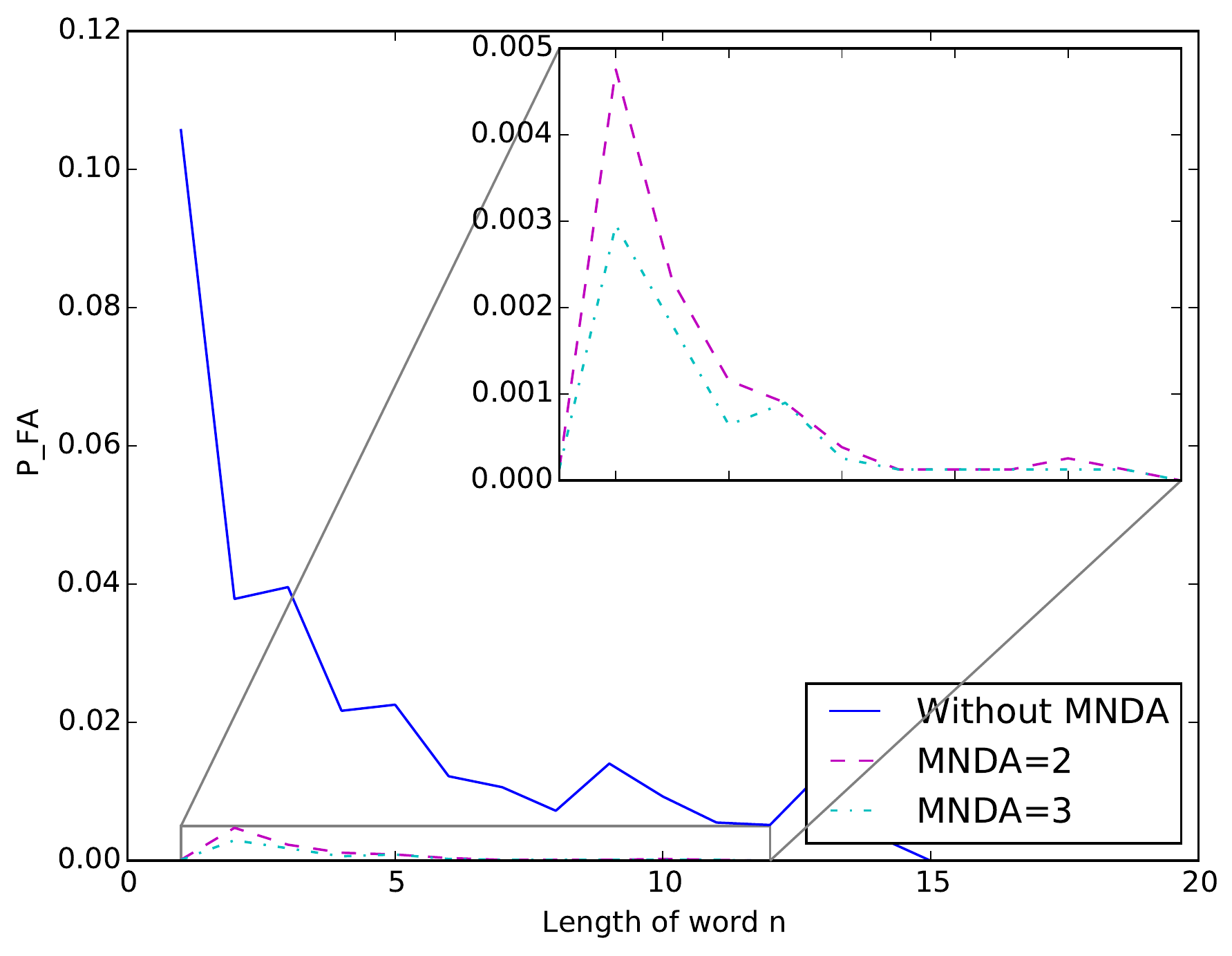}
\caption{Estimated $P_{FA}$ over the length of word $n$ for a given LSTM recognizer trained on the Rimes dataset with and without Minimum Number of Decision Agreement. The estimated probability decreases significantly with the MNDA.}
\label{Pwmisc}
\end{figure}

Figure \ref{Pwmisc} shows the estimated probability $P_{FA}$ of a single recognizer with respect to word length $n$.
The probability was estimated by taking the results of one network trained on the Rimes dataset.
One can observe that the probability $P_{FA}$ (blue curve) is high for short words (less than 4 characters) and therefore that the verification rule is not reliable enough. 
That is why we introduce the Minimum Number of Decision Agreement (MNDA), that allows to  drastically reduce the probability of false acceptance to a very acceptable level. 
This MNDA is the minimum number of classifiers that must take the same classification decision, among the classifiers that have been activated in the cascade. 
%By setting a minimum number of decision agreements (MNDA) between networks the probability with it can be decreased significantly.
As can be observed on magenta and cyan curves from Figure \ref{Pwmisc}, using a MNDA of 2 and 3 leads to a very low estimated probability (below half a percent), even for short words. 

Let us now analyze the effect of the lexicon verification solely (MNDA = 0) on a word recognition task. 
In this respect, Table \ref{ExpePrelBis} shows the performance obtained on the Rimes dataset using a lexicon free BLSTM recognizer, with and without the simple verification rule. 
We see that adding the verification rule to the BLSTM makes the Word Error Rate (WER) dramatically decreases by 93\%, from $33.63\%$ to only $2.25\%$.
By looking at the remaining confusions, most of the errors are type case, accents and plural errors. 
This first experiment demonstrates the strength of a verification strategy in combination with BLSTM classifiers.

\begin{table}[ht]
\centering
\begin{tabular}{|p{3.5cm}|p{2cm}|p{2cm}|p{2cm}|}
\hline
\bf Network &\bf WRR &\bf WER &\bf WJR \\
\hline
BLSTM  & 66.37 & 33.63 & 0\\
BLSTM + verification  & 66.37 & \textbf{2.25} & 31.38\\
\hline
\end{tabular}
\caption{Word Recognition Rate (WRR), Word Error Rate (WER) and Word rejection Rate (WJR) of a lexicon-free BLSTM recognizer on the Rimes dataset, with and without the lexicon verification strategy. Most of recognition errors are rejected thanks to the verification, strongly reducing the Word Error Rate.}
\label{ExpePrelBis}
\end{table}

We have shown that Lexicon Verification can significantly reduce recognition errors, and that the number of False Acceptance can be reduced by adding a Minimum Number of Decision Agreement.
In the design of a cascade we introduce a rejection rule by combining a lexicon verification operator with Minimum Number of Decision Agreement.
Such an operator will serve as an efficient rule to control the decisions at each stage of the cascade.
In the next section, we investigate how to easily generate complementary LSTM RNN.

\subsection{Generation of complementary LSTM RNN}
\label{sec:train}

There are many ways to generate complementary LSTM neural networks.
The first one is by using different architectures (BLSTM vs MDLSTM, changing the number of layers or neurons, etc.) or using different input features (pixels, Histogram of Gradients, etc.).
However these modifications are costly in terms of design, training time, and limited in number.

Some previous experiments have shown that similar LSTM recurrent neural networks trained with different initial weights can be combined with success \cite{menasri2012a2ia}.
These networks have similar recognition rates, but the connection weights are different and therefore they have some complementarity properties which allow to combine them with success. 
However training many networks starting with different initializations is slow as it takes up to a week to train a single network.

In order to get as much complementary networks as possible with a limited effort considering both time in design and training, the idea lies in controlling the training phase of deep neural networks, and is inspired by the work of Choromanska et al. in \cite{choromanska2015loss}. 
This work makes a parallel between fully-connected neural networks loss function and high degree random polynomials which have a huge number of local minima of same order of magnitude.
The authors conclude that when training a large neural network, reaching a local minimum is nearly similar to reaching the global optimum.
As a consequence, exploring these local minima may be a simple but relevant strategy to obtain a large amount of complementary networks that perform equally well, but with different local properties.
We call the ensemble of networks obtained by this strategy a cohort, and we use the obtained cohort to feed a cascade.

\begin{figure}[ht!]
\centering
\includegraphics[width=11cm]{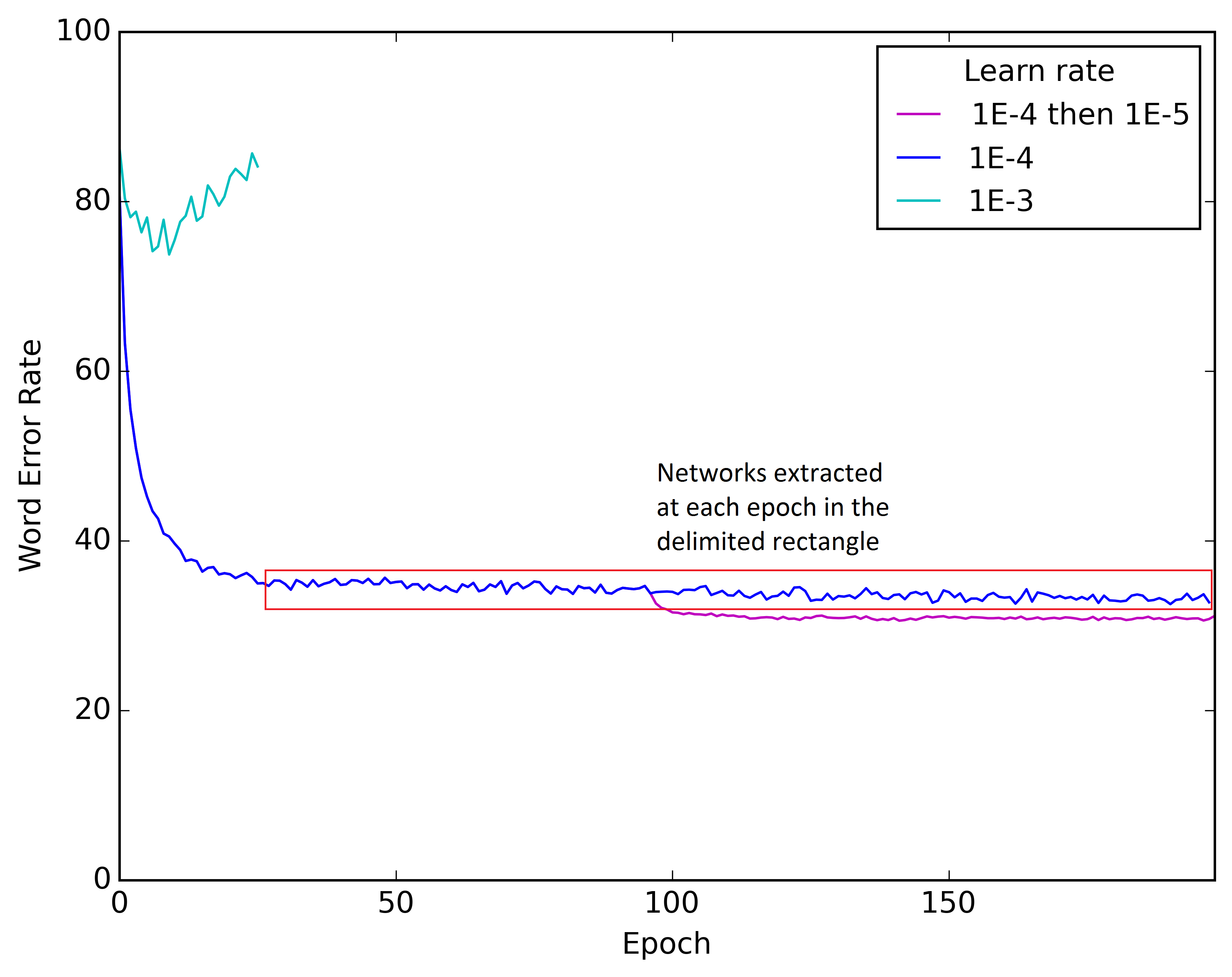}
\caption{Word Error Rate over epoch of a network trained on the Rimes dataset with learning rate: fixed at $10^{-4}$ (blue), at $10^{-4}$ then $10^{-5}$ at epoch 96 (magenta) and at $10^{-3}$ (cyan). On the blue curves there are local fluctuations on the error, when decreasing learn rate on the purple curves fluctuations decrease. On cyan curve the learn rate is too high and training is not converging well.}
\label{curvErr}
\end{figure}

However, this easy and fast strategy to get many complementary networks requires some attention on the training parameters in order to get the desired property.
Indeed training must avoid to be trapped in one local minimum in order to get complementary classifiers.
Training a neural network using steepest gradient descent is controlled by three important parameters which are the learning rate, the momentum value and dataset shuffling.
Momentum is the factor that control the weights update and thus the convergence of the training procedure.
We use a fixed momentum of 0.9, which is commonly used in the literature, in order to help escaping local minima.
When training networks, the examples are shuffled between each epoch in order to get a better convergence by preventing cycles.
As training examples are randomly shuffled between epochs, we can consider that the state of the network (the weights' value) is randomly reached from one epoch to another.
The last and maybe most important parameter is the learning rate.
It is the weight associated to the value of the gradient that serve as the update rule of the network weights.
Figure \ref{curvErr} shows the impact of the learning rate on training a BLSTM network defined in \ref{sec:archi} on the Rimes dataset (see \ref{sec:dataEval}).
A large learning rate does not allow convergence of the network, as shown on Figure \ref{curvErr} (cyan curve) where the network does not converge properly with a learning rate of $10^{-3}$.
A too small learning rate leads to very small changes of the weights values and does not allow reaching another local minimum, and as a consequence gives no complementary networks between epochs. 
As shown on Figure \ref{curvErr} on the magenta curves decreasing the learning rate to $10^{-5}$ at the epoch 96 (taken arbitrarily but any other epoch at this stage would have been suited) produces less fluctuation of the WER leading to less variability and then nearly no complementarity.
The desired behavior is obtained for a learning rate set to $10^{-4}$.

By choosing these parameters, the networks extracted at each epoch of one single complete training phase can constitute a cohort.
The red rectangle on Figure \ref{curvErr} shows the region where the networks can be selected, where training has converged to reach a region with no overall significant decrease of the WER, but with small fluctuations that highlight the various local minima reached, thus providing networks with different local properties. 
The proposed training strategy although simple is very effective to get complementary networks.
Some experimental analysis are presented in section \ref{sec:analComp} to evaluate the network complementarity within a cohort.

At this stage, we have proposed a lexicon verification rule that can serve as an effective reliable rejection rule, and an easy mean to get hundreds of complementary LSTMM RNN during a single training session.
Let us now describe the implementation details of the cascade.

\section{Implementation and Results}

Based on the methodological principles established in the previous section, we now detail our implementation and present the results achieving state of the art performance. 
We also analyze the method regarding both the complementarity between networks and the lexicon verification stage in terms of performance.

\subsection{Dataset and Evaluation}
\label{sec:dataEval}
The experiments have been carried out on the Rimes\cite{grosicki2009icdar} and IAM \cite{marti2002iam} isolated words datasets and using the official splits and lexicons provided with these datasets.
The character sets contain upper and lower case characters of the Latin alphabet, symbols that may occur in a word like "\textbf{'}" or "\textbf{-}", and many others.
For the Rimes dataset there is also accented characters, and the evaluation is made on lower case as in \cite{menasri2012a2ia} for comparison purpose, and also because of some ground truth ambiguity \footnote{Namely on certain upper case characters, especially for letter "\textbf{j}" in the word "\textbf{je}" or "\textbf{j'}" where many ground truth errors occur in the dataset.}.

The lexicons for both datasets are composed of all the words of the training, validation and test sets, as used in the competitions and related papers, giving lexicons size of 5744 and 12202 words for Rimes and IAM respectively. 
For the Rimes dataset, a smaller lexicon composed of test words (1692) is used in order to evaluate lexicon sensitivity. 
Notice that for every experiment, the lexicons have a 100\% coverage rate as it is the case for every evaluation conducted for related word recognition in the literature.

We also propose to evaluate the performance of our approach on very large lexicons that better represent what can be found in industrial applications (e.g. It exists millions of named entities for addresses : countries, regions, towns, street names, names,...). 
For that, we have collected by ourselves two extremely large lexicons for each dataset\footnote{These lexicons are available on request.}:
\begin{itemize}
\item The first one is an union of the words from the french Wikipedia, Wiktionnaire, French dictionary Gutemberg and the Rimes dataset, leading to French lexicon of \textbf{3,276,994} words;
\item The second one is the union of the words found in the one billion word data \cite{chelba2013one} and the IAM dataset, leading to an English lexicon of \textbf{2,439,432} words.
\end{itemize}

For all the experiments, we measure the performance with the Character Error Rate (CER), calculated with the Levenshtein distance, and the Word Error Rate (WER).
We also measure the Word Recognition Rate (WRR) and the Word Rejection Rate (WJR).
In the classifier cascade only, character errors of rejected words are not considered in the CER.

\subsection{Architecture}
\label{sec:archi}
Two well established LSTM RNN architectures have been used for the experimentations:

The first is a BLSTM architecture identical to the one used in \cite{mioulet2015exploring}: it is a two layers network composed respectively of 70 and 120 LSTM blocks, separated by a subsampling layer of 100 hidden neurons without bias, and with an hyperbolic tangent activation function.
The network also has two layers to reduce the sequence length.
The first one concatenates the input vectors in pairs, while the second one concatenates the output vectors of the first layer in pairs.
For example, a sequence composed of 12 frames of 1 pixel width is transformed into a sequence of length 3 corresponding to 3 frames of 4 pixels width.
As input features, we use histogram of oriented gradients (HOG) \cite{dalal2005histograms} that have demonstrated their efficiency for handwriting recognition \cite{Bid15}. 
Images are normalized to 64 pixels height.
A sliding window of 8 pixel width extracts the HOG features at a 1 pixel pace.
This architecture has been selected for its balanced characteristics, both performing slightly better than the reference architecture (presented below) \cite{graves2009offline} and allowing a fast decoding, 15 milliseconds per word on average (intel i7-3740QM CPU).
%Other experiments showed us that networks with same order of parameters, have similar performances.

The second architecture is a MDLSTM architecture similar to the reference in \cite{graves2009offline}. 
This architecture is composed of three layers of respectively 2, 10 and 50 LSTM cells, and two subsampling layers of 6 and 20 neurons. 
Raw images are normalized and directly fed to the network.
The decoding time is 10 milliseconds per word on average (on an intel i7-3740QM CPU).

For both architectures we use a lexicon free "Best path decoding" algorithm \cite{graves2012supervised} described in (\ref{sec:lstm}) to retrieve the characters string.
Training and decoding have been performed with RNNLIB \cite{graves2013rnnlib}.

\subsection{Generating a cohort of LSTM RNN}
\label{sec:cohorts}

To get a cohort of complementary LSTM RNN, the three following strategies are followed:
(i) Our training trick described in section \ref{sec:train}, for which we get one network per epoch ;
(ii) Different starting initialization ;
(iii) Two different architectures BLSTM and MDLSTM.

\begin{figure}[ht!]
\centering
\includegraphics[width=11cm]{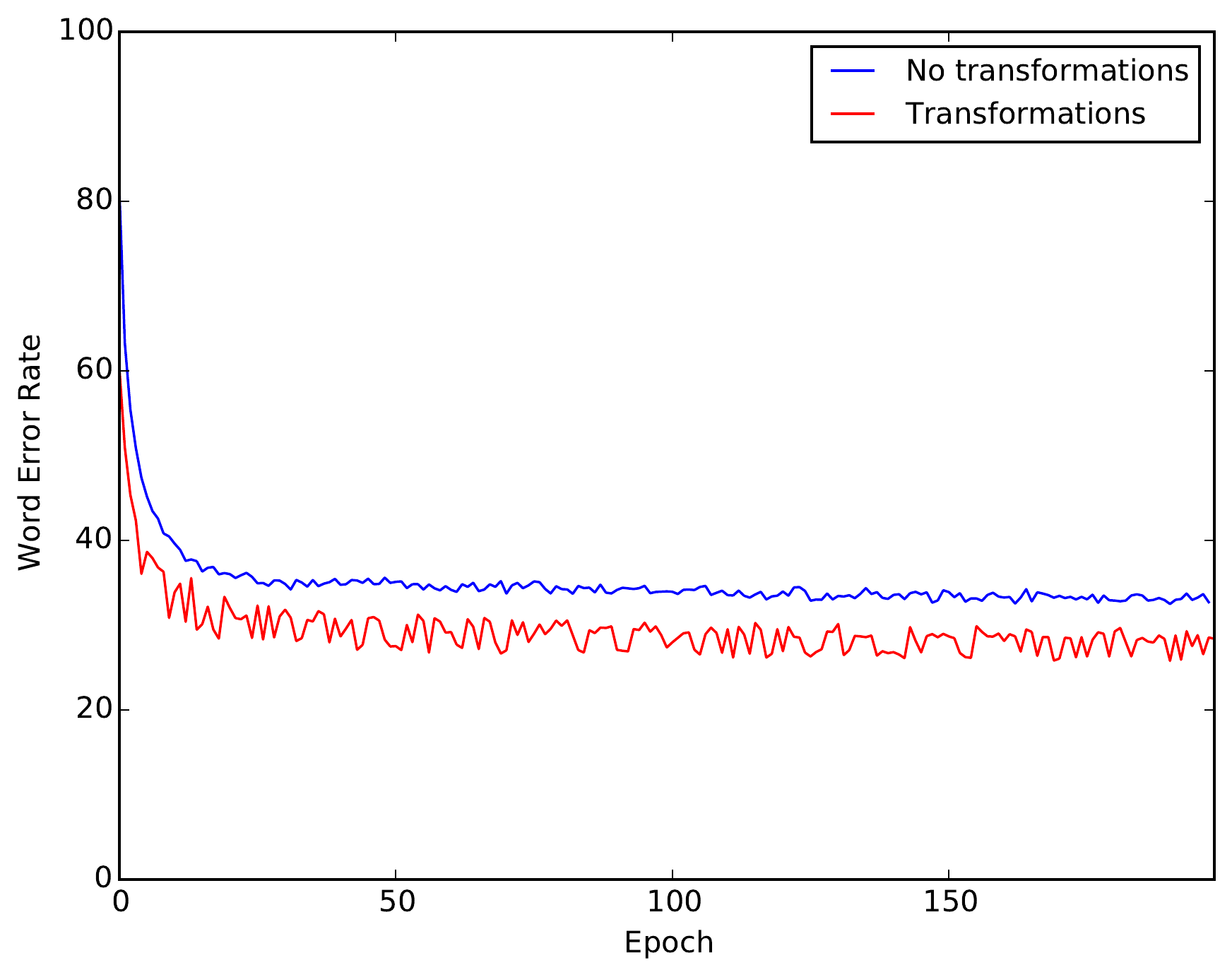}
\caption{Comparisons of Word Error Rate over of two networks with and without transformations with learn rate fixed at $10^{-4}$. The error fluctuates at a higher amplitude for the network with transformations.}
\label{curvErrDef}
\end{figure}

Moreover, we use a common trick to improve performance of neural networks by performing transformations over the training set, in order to increase the size of the training set. 
By using rotations and warping, we multiply the size of the training set by 3, and as shown in Figure \ref{curvErrDef} the word error rate improves. 
Beside providing a lower error, we observe that the transformations also bring more fluctuations during training by adding more training samples between two epoch, thus increasing the number of weight updates and the potential local differences between two network between two epochs. 

2100 different networks are collected from only ten different trainings on the Rimes dataset with different random initializations:
\begin{itemize}
\item 2 BLSTM trainings;
\item 1 MDLSTM training;
\item 4 BLSTM trainings with transformations;
\item 3 MDLSTM trainings with transformations.
\end{itemize}
Training these 10 networks with 3 computers on CPU only took 10 days, while training 2100 networks with different initialization would have taken 2 years and 9 months.
 
Regarding the IAM dataset, we also trained two networks with transformations (tripling the size of train set), two BLSTM and two MDLSTM, leading to a total of 1039 networks for this task.

The purpose of doing several training is threefold, it allows us to: i) get more networks with training run in parallel, (ii) check successfully that the method doesn't depend on the initial weights or architecture selected, and (iii) improve performance thanks to the two architectures and using different features.

\subsection{Results}

\subsubsection{Performance of cascades built from a single cohort}

In this section, cascades are built using networks from a single cohort of RNN.

Cascade from single cohort of BLSTM, BLSTM with transformations, and  MDLSTM are considered. 
We select 100 networks to form each cohort and combine each cohort in a cascade, using the lexicon verification operator.

When combining hundreds of networks into a cascade, the performance depends on ordering the classifiers in the cascade, and on the value of the Minimum Number of Decision Agreements (MNDA) in the decision stage. 
Ordering the classifiers in the cascade is made according to the deletion error criterion (deletion is the rate of characters omitted by the network) on the validation set.
Tuning the MNDA parameter was carried out by considering long words and short words.

As previously seen in Figure \ref{Pwmisc}, we observe that short words are more prone to mistakes, justifying the fact that short and long words will be considered differently in the rejection stage of the cascade.
In these experiments we have chosen a MNDA of 3 for long words and 10 for short words, however, small deviations from the values do not affect the results much.
For example when using MNDA in the range 2 to 10 for long words, and MNDA in the range 5 to 40, we observe no modification in the CER, while WER is modified by 0.25\% only.
It is also possible to optimize these parameters regarding the validation set.

In Table \ref{tab:resOneCohort}, the first line is a recall of the performance of a single BLSTM obtained in our preliminary experiments (See section \ref{sec:verif} ).
A Word Recognition Rate of 66\% is obtained. 
When using transformations, this score improves to 71\%
Combining 100 networks highly improves the WRR up to 84.5\%, while the same experiment with transformations and 100 BLSTM improves the WRR to 89.56\%.
Finally, it appears that improvement does not depend on the type of network nor on the input features, as we observe using MDLSTM a significant improvement of the WRR, on the last line of the Table \ref{tab:resOneCohort}.

\small{
 \begin{table}[ht!]
\centering
\begin{tabular}{|p{6.8cm}|M{0.9cm}|M{0.9cm}|M{0.9cm}|M{0.9cm}|}
\hline
\textbf{System}  & \textbf{WRR} & \textbf{WER} & \textbf{WJR} & \textbf{CER} \tabularnewline
\Xhline{0.5mm}
Single BLSTM  & 66.37 & 33.63 & 0 & 11.24 \tabularnewline
Single BLSTM (+transformations) & 71.78 & 28.22 & -  & 8.89\tabularnewline
Cascade of 100 BLSTM  & 84.53 & 3.13 & 12.34 & 0.99 \tabularnewline
Cascade of 100 BLSTM (+transformations) & 89.56 & 2.74 & 7.70 & 0.82 \tabularnewline
Cascade of 100 MDLSTM & 80.27 & 4.09 & 15.64 & 1.40 \tabularnewline
\hline
\end{tabular}
\caption{Word Recognition Rate, Word Error Rate, Word reJection Rate and Character Error Rate of different Systems on the Rimes dataset with the 5744 words lexicon. Networks from the cascade are extracted from a single training procedure (i.e. cohort).}
\label{tab:resOneCohort}
\end{table}}
 
We now analyze the results for cascades made of classifiers from multiple cohorts, in the hope to further increase the complementarity within the cascade.

\subsubsection{Performance of cascades with multiple cohorts}

We evaluate the cascade of cohorts of LSTM for the Rimes and IAM datasets.
First we report the results on the Rimes dataset. 
By gathering the 10 cohorts described in section \ref{sec:cohorts}, 2100 networks are combined in the cascade.  
As shown in Table \ref{tab:resRimes}, we achieve state of the art performance for a 5744 words lexicon on the Rimes dataset.
To compare the results with others, which have no reject, we use a Viterbi lexicon decoding on the rejected words at the end of the cascade.
For doing so, character posterior probabilities are the average class posterior probabilities of ten LSTM networks randomly picked among the cohort, before running the Viterbi decoder.
Averaging the output probabilities from different networks improves the performance (in comparison to taking the output of only one network). Notice that the number of networks randomly selected has a very low impact on the performance above a certain threshold. 
Selecting 5, 10, 20 or 40 networks yields nearly similar results.
The gap between this study and the results presented in \cite{poznanski2016cnn}, both in terms of CER and WER is significant with an absolute decrease of 0.56 (29\%) of the CER and 0.42 (11\%) of the WER. 

The recognition difference between small (1692 words) and large (5744 words) lexicons is very small ($\simeq$ 0.48 points) compared to the traditional performance drop observed. 
The difference between the large (5744 words) and gigantic (> 3M words) lexicon is more important, however it is still rather limited with a difference of 5.71\%.
Both results prove the low sensitivity of the method to the size of the lexicon.

Regarding the gigantic lexicon, the system provides very interesting recognition rate (90.14\%) as the lexicon is nearly 600 times larger than the Rimes lexicon. Such lexicon size opens new perspectives for processing named entities or multilingual documents for example.
Notice that we have no comparison with any other study since to the best of our knowledge, there is no other reference in the literature using such lexicon size. 

\small{
 \begin{table}[ht!]
\centering
\begin{tabular}{|M{3.5cm}|M{2.2cm}|M{1cm}|M{1cm}|M{1cm}|M{1cm}|}
\hline
\textbf{System} & \textbf{Lexicon size} & \textbf{WRR} & \textbf{WER} & \textbf{WJR} & \textbf{CER} \tabularnewline
\Xhline{0.5mm}
This work & \textit{1692} & 96.08 & 2.32 & 1.60 & 0.76 \tabularnewline
This work & \textit{5744} & 95.60 & \textbf{3.00} & 1.40 & 0.99 \tabularnewline
This work & \textit{3 276 994} & 90.14 & 9.18 & 0.68 & 2.67 \tabularnewline
This work + Viterbi & \textit{5744} & \textbf{96.52} & 3.48 & - & \textbf{1.34} \tabularnewline
\hline
Menasri et al. \cite{menasri2012a2ia} & \textit{5744} &95.25 & 4.75 & - & -  \tabularnewline
Poznanski et al. \cite{poznanski2016cnn} & \textit{5744} &96.10 & 3.90 & - & 1.90  \tabularnewline
\hline
\end{tabular}
\caption{Results of the 2100 networks cascade on the Rimes dataset. The cascade achieves state of the art performance.(WRR: Word Recognition Rate, WER: Word Error Rate, WJR: Word reJection Rate, CER: Character Error Rate)}
\label{tab:resRimes}
\end{table}}

The results on the IAM dataset are presented in Table \ref{tab:resIam}.
Both CER and WER are better than state of the art ones by respectively 0.66 (19\% decrease) and 0.52 (8\% decrease).
The result for the gigantic lexicon is also impressive with 85\% of words recognized with a lexicon 200 times larger than the initial one.

 \begin{table}[ht!]
\centering
\begin{tabular}{|M{3.5cm}|M{2.2cm}|M{1cm}|M{1cm}|M{1cm}|M{1cm}|}
\hline
\textbf{System} & \textbf{Lexicon size} & \textbf{WRR} & \textbf{WER} & \textbf{WJR} & \textbf{CER} \tabularnewline
\Xhline{0.5mm}
This work & \textit{12202} & 92.46 & \textbf{5.04} & 2.50 & 2.08 \tabularnewline
This work & \textit{2 439 432} & 85.51 & 13.30 & 1.19 & 4.77 \tabularnewline
This work + Viterbi &  \textit{12202} & \textbf{94.07} & 5.93 & - & \textbf{2.78} \tabularnewline
\hline
Poznanski et al. \cite{poznanski2016cnn} & \textit{12202} & 93.55 & 6.45 & - & 3.44  \tabularnewline
\hline
\end{tabular}
\caption{Results of the 1039 networks cascade on the IAM dataset. The cascade achieves state of the art performance. (WRR: Word Recognition Rate, WER: Word Error Rate, WJR: Word reJection Rate, CER: Character Error Rate)}
\label{tab:resIam}
\end{table}

%For both datasets our approach achieves state of the art results, and it has a low sensitivity regarding the lexicon size and can deal with gigantic lexicons.

\subsection{Analysis of the networks complementarity}
\label{sec:analComp}

\begin{figure}[ht!]
\centering
\includegraphics[width=11cm]{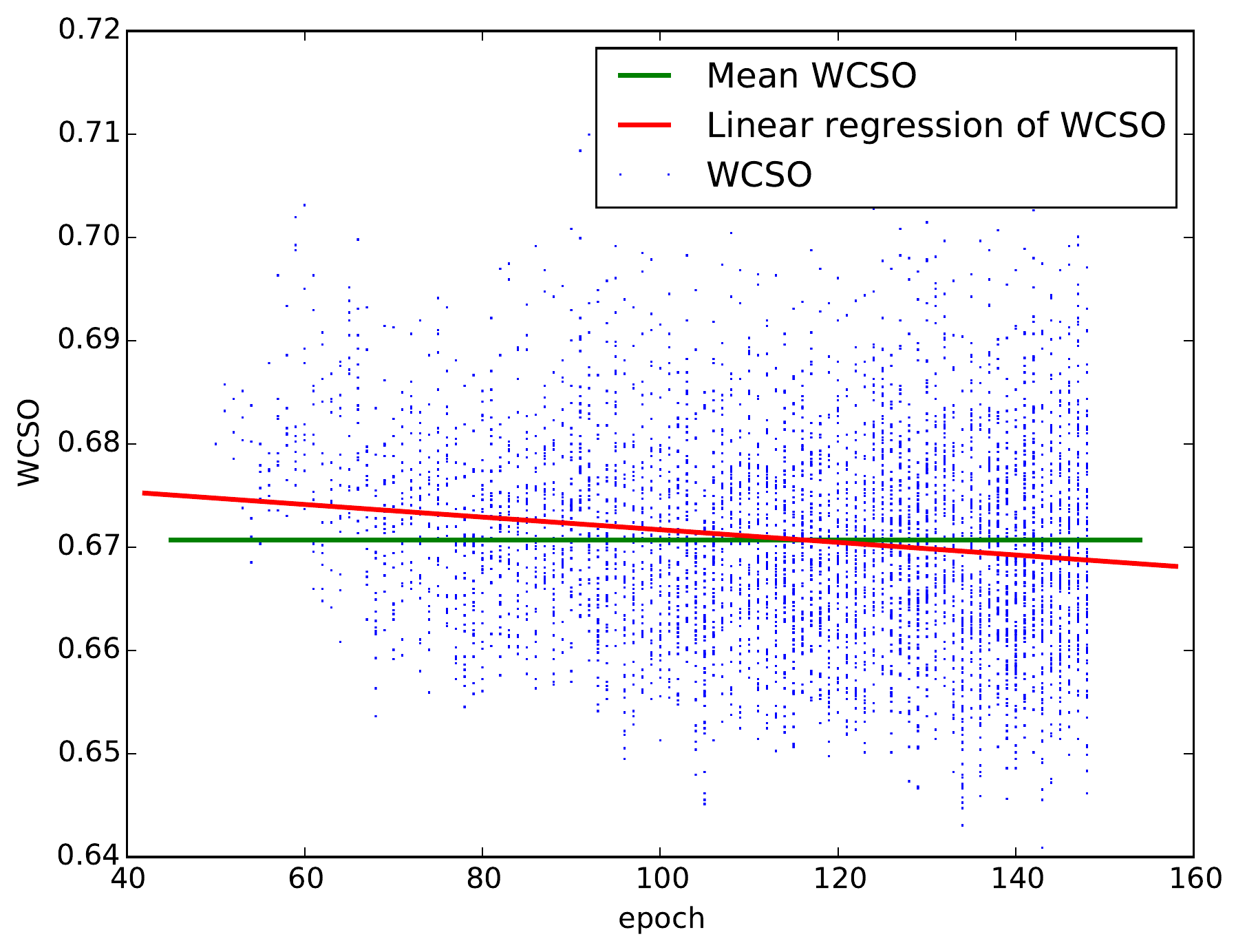}
\caption{Word Classifier Similarity Outputs over the epoch of training (blue dots) with the mean WCSO (green curve) and the linear regression of WCSO (red curve). }
\label{curvSim}
\end{figure}

In this subsection, we analyze the complementarity of the classifiers selected in the cascades. 
For this purpose, we estimate the Word Classifier Similarity Outputs (WCSO) of every networks during the training phase.
WCSO measures the percentage of identical answers at word level between two classifiers.
We compute the WCSO for 100 networks taken during the epochs as shown on Figure \ref{curvErr}. The results are presented on Figure \ref{curvSim}.

We measured an average similarity of 67\% between two networks with a standard deviation of only 0.009, proving the complementarity of the networks selected by the proposed procedure(whereas for a learning rate of $10^{-5}$ the mean similarity is 90\%).
As shown on Figure \ref{curvSim}, the linear regression of the WCSO is decreasing, meaning that the complementarity between networks is increasing with the number of epochs.

\begin{figure}[ht!]
\centering
\includegraphics[width=11cm]{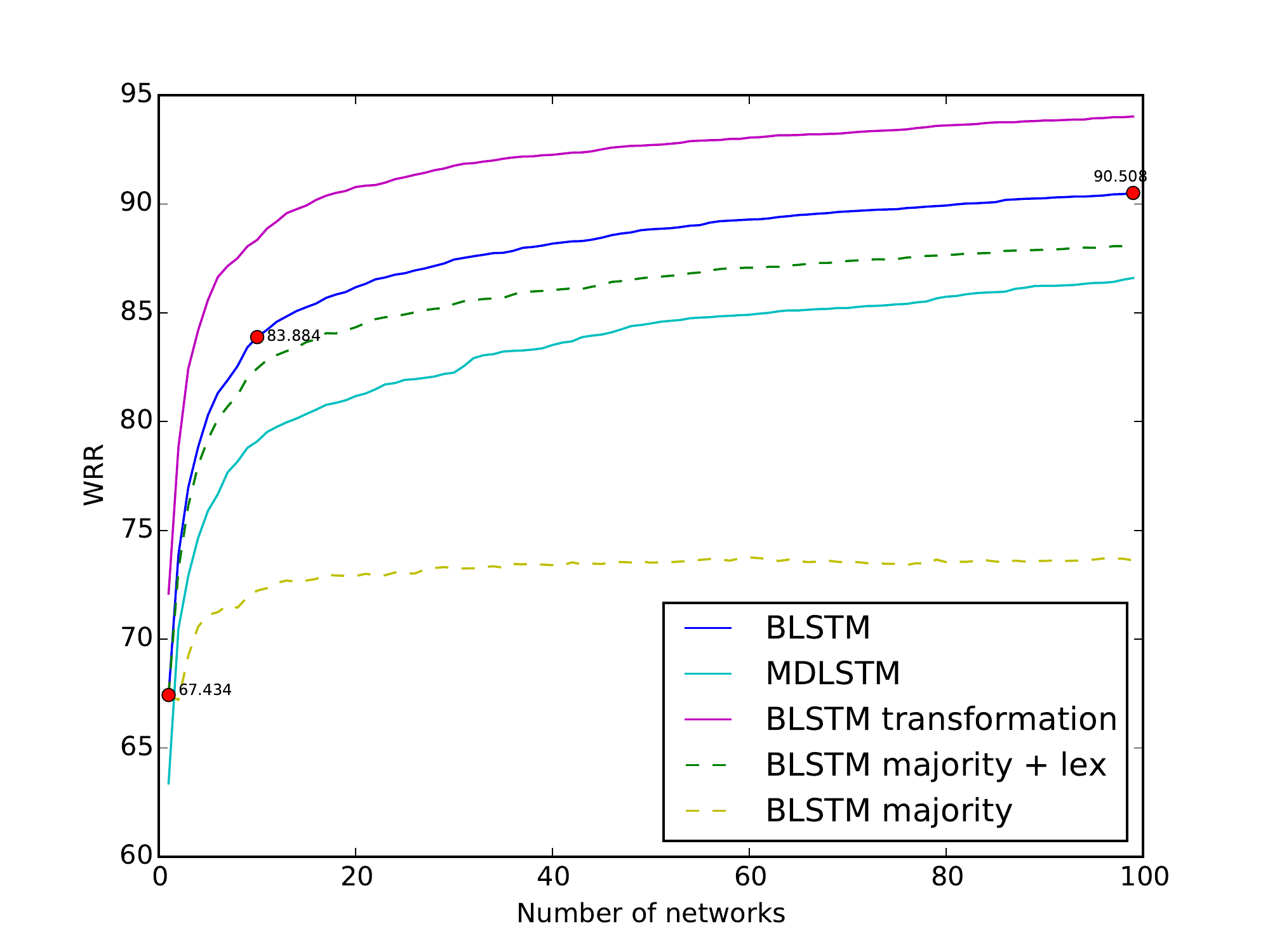}
\caption{Evolution of the optimal theoretical recognition over the number of networks for three different cohort. The theoretical optimal recognition is improving with the cohort size.}
\label{curvWr0}
\end{figure}

To further assess the complementarity of the networks, we compute the percentage of having at least one correct hypothesis among the cohort. 
%That is to say we took the outputs of $n$ networks and if the right result is among the output then we consider it as recognized.
Figure \ref{curvWr0} shows this recognition over the number of networks and has been computed for three different cohort from a BLSTM, a MDLSTM and a second BLSTM with transformations (the same trainings that were presented in Table \ref{tab:resOneCohort}).
For one BLSTM network (blue curve) this percentage is around 66\% whereas for 10 networks it is around 83\% and for the 100 BLSTM it is about 90\%.
The positive evolution of the word recognition rate without any stabilization on Figure \ref{curvWr0} for each cohort shows that the more networks from a cohort, the more likely the right solution can be found.

Moreover, Figure \ref{curvWr0} also shows the results for the networks from the BLSTM cohort using two simple decision rules: a majority vote and a majority vote with lexicon verification.
The majority vote selects the most frequent word hypothesis, whereas the majority vote with lexicon verification selects the most frequent word hypothesis that belongs to a lexicon.
Figure \ref{curvWr0} shows that the simple majority vote provides limited word recognition rate, whereas when the lexicon verification improves the performance of more than 10\%.
This illustrates the interesting capacities of the lexicon verification and its reliability.

\subsection{Processing time}

We analyze now one potential drawback of the approach, which requires many networks and thus a large amount of processing time.
However, the cascade architecture is very effective regarding computation cost, because once a candidate is accepted by the verification stage, it does not pass through the rest of the cascade.
Regarding the processing time due to the lexicon verification, it is below one microsecond, even considering the gigantic lexicons presented in the previous section.
Finally, when considering our system with 2100 networks, the mean processing time per word is 729 milliseconds. 80\% of the words are processed in less than 0.175s, after 14 networks or less, and 90\% in less than half a second (after 41 networks or less)\footnote{This timing has been evaluated on an Intel CPU i7-3740QM.}.
As our code has not been optimized nor parallelized, we believe that there is room for strong  improvements.

Considering memory issues, the networks involved in these experiments do not require a lot of memory space, as the whole set 2100 networks fit into 6GB, even with double precision encoding (64 bits).

Above timing consideration, one would ask whether the whole cohort of networks is necessary, or if a subset of them would perform similarly, which in this case would also alleviate the processing time. This point is addressed in the next section.

\subsection{Pruning classifiers from the cohort}

The aim of the selection is to decimate the cohort of classifiers so as to keep the most efficient reduced set of classifiers.
This has been achieved by simply removing the networks providing the poorest performance on the validation dataset.
Networks which don't recognize new words, or which yield too much false acceptance are removed.
By doing so, we manage to reduce the number of networks from 2100 to 118.
As shown in Table \ref{tab:resPruning}, we observed a very acceptable performance drop compared to the cohort of 2100 networks. 
Nearly similar results are obtained when using Viterbi decoding of the rejects: the WER is \textbf{3.64\%} (previously 3.48\%) and the CER is \textbf{1.49\%} (previously 1.34\%). 
Note that this reduced architecture still improves state of the art. 
The mean processing time of this pruned cohort is decreased to 197 milliseconds (previously 729ms).

\begin{table}[ht!]
\centering
\begin{tabular}{|M{3.5cm}|M{2.2cm}|M{1cm}|M{1cm}|M{1cm}|M{1cm}|}
\hline
\textbf{System} & \textbf{Lexicon size} & \textbf{WRR} & \textbf{WER} & \textbf{WJR} & \textbf{CER} \tabularnewline
\Xhline{0.5mm}
This work & \textit{5744} & 92.96 & 2.28 & 4.76 & 0.69 \tabularnewline
This work & \textit{3 276 994} & 88.82 & 8.04 & 3.14 & 2.20 \tabularnewline
This work + Viterbi & \textit{5744} & 96.36 & 3.64 & - & 1.49 \tabularnewline
\hline
\end{tabular}
\caption{Results of the pruned cascade of 118 networks on the Rimes dataset. Results are close to our previous results with 18 times less networks.}
\label{tab:resPruning}
\end{table}

\section{Conclusion}

This works presents a new recognition paradigm that substitute the traditional lexicon directed recognition by a lexicon verification procedure. It is exploited in a cascade framework involving hundreds of LSTM recurrent neural networks.
The networks are obtained during a single training procedure, therefore not requiring many parameter optimization and a limited training duration. 
By analyzing the Word Classifier Similarity Outputs on the Rimes datasets, we prove that complementarity can be obtained by this procedure and is decreasing with epochs.

Another important part of the success of the cascade combination is due to the lexicon verification operator including both the lexicon verification and the Minimum Number of Decision Agreement.
The combination shows a low probability of false acceptance, and has also very low sensitivity to lexicon size.
Besides, our method allows to process gigantic lexicons without extra processing time, which has never been done before.
Even if the performance decreases with the lexicon size, MNDA could be increased with the lexicon size in order to reduce the generated confusions.
We successfully achieve state of the art results for both Rimes and IAM datasets, with nearly no parameters to tune. Moreover, the system can be customized for application's needs.

One interesting perspective to this work is to extend the idea at line level by using both the cohort and the lexicon verification with a line level suited combination framework as ROVER.
As the lexicon size is not an issue anymore, it also opens new perspectives for large lexicon applications such as the processing of multilingual documents, or named entity recognition.

%The way we obtain our cohort of networks and the results we get raises some interrogations, especially about the current trend of deeper and deeper architectures. 
%Is covering a huge number of local minimum with simple networks and combined their results in simple ways working for every application ? 
From a methodological point of view, it would be interesting to investigate the genericity of the cohort generation, and especially to analyze its efficiency using alternative architectures like convolutional neural networks or adversarial neural networks.
%Extending our cohort principle to other neural networks and applications, like image description seems to be a promising path to explore.

\section*{References}

\bibliography{biblio}

\end{document}